\documentclass{article}
\usepackage{spconf,amsmath,graphicx}

\makeatletter
\let\NAT@parse\undefined
\makeatother

\usepackage{graphics,mathptmx,times,amssymb}
\usepackage{amsmath,epsfig,xcolor,graphicx,url,xspace,booktabs}
\usepackage{url,multirow,bm,bbm,breqn,rotating,color,colortbl,subcaption,pifont,colortbl}

\usepackage{enumitem}

\definecolor{Gray}{gray}{0.86}

\usepackage{hyperref}
\definecolor{citecolor}{HTML}{0071BC}
\definecolor{linkcolor}{HTML}{ED1C24}
\definecolor{Gray}{gray}{0.86}

\hypersetup{colorlinks=True, citecolor=citecolor, urlcolor=magenta}

\title{Deep EEG Super-resolution: Upsampling EEG Spatial Resolution with Generative Adversarial Networks}

\name{Isaac Corley\sthanks{Corresponding author \tt\footnotesize 
\href{mailto:isaac.corley@utsa.edu}{isaac.corley@utsa.edu}}, Yufei Huang}
\address{University of Texas at San Antonio}

\begin{document}

\maketitle
\thispagestyle{empty}
\pagestyle{empty}

\begin{abstract}
\textbf{Electroencephalography (EEG) activity contains a wealth of information about what is happening within the human brain. Recording more of this data has the potential to unlock endless future applications. However, the cost of EEG hardware is increasingly expensive based upon the number of EEG channels being recorded simultaneously. We combat this problem in this paper by proposing a novel deep EEG super-resolution (SR) approach based on Generative Adversarial Networks (GANs). This approach can produce high spatial resolution EEG data from low resolution samples, by generating channel-wise upsampled data to effectively interpolate numerous missing channels, thus reducing the need for expensive EEG equipment. We tested the performance using an EEG dataset from a mental imagery task. Our proposed GAN model provided 10^4 fold and 10^2 fold reduction in mean-squared error (MSE) and mean-absolute error (MAE), respectively, over the baseline bicubic interpolation method. We further validate our method by training a classifier on the original classification task, which displayed minimal loss in accuracy while using the super-resolved data. The proposed SR EEG by GAN is a promising approach to improve the spatial resolution of low density EEG headsets.}
\end{abstract}

\section{Introduction}
Electroencephalography (EEG) is a noninvasive neuroimaging modality widely used for clinical diagnosis of seizures and cognitive neuroscience. It has gained increasing popularity in recent years as a neurofeedback device in   brain-computer interface (BCI) systems with applications including typing interface for locked-in patients, neurorehabilitation~\cite{wolpaw1994multichannel}, brain-controlled drone~\cite{merino2017asynchronous}, and detection of driver fatigue~\cite{hajinoroozi2016eeg}. However, a primary bottleneck to EEG-based BCI research is the cost of hardware. Ideally, EEG devices with high-density channels are preferred in order to obtain recordings of brain activities with high spatial resolution underlying different cognitive events. However, the cost of EEG hardware increases exponentially with channels, with a majority of commercial EEG devices with 32 channels costing more than \$20k. For academia and industry, this can greatly hinder the quality of products and research being performed. This also results in poor generality for EEG-based algorithms because prediction algorithms developed using one headset cannot be used for a headset of different channels even if both are used to measure the same cognitive events. EEG channel interpolation has been proposed in many research efforts~\cite{petrichella2016channel, courellis2016eeg} to recreate missing or defective sensor channels. Although they showed favorable improvement for single selective channel interpolation, research for interpolating many channels at a global scale is scarce.  

Deep learning and its applications have recently become highly popular and rightly so, due to their superior ability to learn representations of complex data~\cite{lecun2015deep}. One of its applications is image super-resolution (SR), where deep learning-based pixel interpolation was developed~\cite{dong2015image} to generate high-resolution (HR) copies from low-resolution (LR) images. The state-of-the-art SR performance is obtained by the new game theoretic deep generative model of Generative Adversarial Networks (GANs)~\cite{goodfellow2014generative}, which established the first framework to achieve photo-realistic natural images for a 4x upscaling factor~\cite{ledig2017photo}. 

Inspired by the similarity between global EEG channel interpolation and image SR, along with the superb image SR performance achieved by GANs, we propose deep EEG super-resolution, a novel framework for generating high spatial resolution EEG data from low resolution recordings using GANs. We compare our work to a baseline of bicubic interpolation and then additionally verify performance by training a classifier using the SR EEG data for the classification purpose of the original EEG dataset. 

\section{Data}
\subsection{Berlin BCI Competition III, Dataset V}
Dataset V of the Berlin Brain Computer Interface Competition III provided by the IDIAP Research Institute~\cite{millan2004need} was used. The dataset consists of 32 EEG channels recorded at 512 Hz for 3 individual subjects, located at the standard positions of the International 10-20 system \textit{(Fp1, AF3, F7, F3, FC1, FC5, T7, C3, CP1, CP5, P7, P3, Pz, PO3, O1, Oz, O2, PO4, P4, P8, CP6, CP2, C4, T8, FC6, FC2, F4, F8, AF4, Fp2, Fz, Cz)}. The dataset was initially used for a mental imagery multiclass classification competition of three labeled tasks. The data is provided with train and test sets for each subject; however, only the train set was used due to the test set not being provided with labels. The train set contained a total of 1.1M samples. 

\subsection{Preprocessing}
Following the epoch extraction procedure in~\cite{millan2004need}, the raw data was separated into epochs of length 512 samples using a moving window with a stride of 32 samples. This resulted in epochs of size (32 channels by 512 samples). The epoched data was then split into train, validation, and test sets for holdout validation using a 75/20/5 percentage split criterion.  

The initial Dataset V classifiers used precomputed features, which consisted of the estimated power spectral density (PSD) of each epoch in the band 8-30 Hz with a frequency resolution of 2 Hz for the 8 centro-parietal channels \textit{(C3, Cz, C4, CP1, CP2, P3, Pz, P4)}, which resulted in a 96-dimensional vector (8 channels, 12 frequency components). 

For use with super-resolution models, all datasets were reshaped to epochs of size (32 channels by 64 samples). To produce the low-resolution (LR) data, the epochs were downsampled by channel based upon the scale factor used, e.g., downsampling 32 channels by a scale factor of 2 would remove every other channel, leaving 16 channels of LR data. The removed channels are then used as the HR data. The input data and its corresponding ground truth were then standard-normalized to a mean $\mu=0$ and standard deviation $\sigma=1$ using the mean and standard deviation of the input channel training set. This was repeated using the same statistics for normalizing the validation and test data. 

\section{Methods}
\begin{figure}[ht!]
    \centering
    \includegraphics[width=0.65\linewidth]{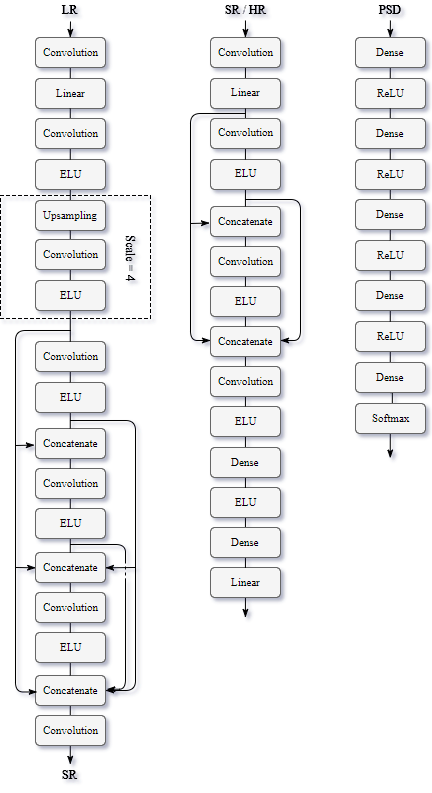}
    \caption{\textbf{Model Architectures.} \textit{Generator (left), Discriminator (center), Classifier (right)}}
    \label{fig:model}
\end{figure}

\subsection{Generative Adversarial Networks}
Generative Adversarial Networks (GANs) are an unsupervised deep learning framework recently proposed by Goodfellow et al.~\cite{goodfellow2014generative}. The framework is composed of two networks, a generator G and a discriminator D, optimized to minimize a two-player minimax game, where the generator learns to fool the discriminator and the discriminator learns to prevent itself from being fooled. As Goodfellow et al.~\cite{goodfellow2014generative} describes, \textit{"The generative model can be thought of as analogous to a team of counterfeiters, trying to produce fake currency and use it without detection, while the discriminative model is analogous to the police, trying to detect the counterfeit currency."} During training of GANs, the generator is fed an input noise vector and produces an output distribution $P_G$. The discriminator is then trained to learn to discriminate between $P_G$ and the true data distribution, $P_{Data}$. Additionally, the generator is trained to learn how to further fool the discriminator. Theoretically, $P_G$ will converge to $P_{Data}$ with the discriminator being unable to differentiate between generated and true samples, resulting in an ideal generative model which can produce data following the true data distribution. 

While GANs are a powerful framework, they possess stability issues which cause the adversarial networks to rarely reach convergence. Variations of the framework, namely Wasserstein GANs (WGANs)~\cite{arjovsky2017wasserstein, gulrajani2017improved}, have been developed which use different loss functions with properties that improve training stability. In contrast to the original GAN framework, WGANs minimize the Earth Mover’s Distance (Wasserstein-1 Distance) and attempt to constrain the gradient norm of the discriminator’s output with respect to its input using a gradient penalty in the loss function. We adopt the WGAN framework for training throughout our research as we experienced improved stability over the original GAN framework. 

\subsection{Proposed Wasserstein Generative Adversarial Networks for EEG Super Resolution}

\begin{table}[ht!]
\centering
\caption{\textbf{Generator Model Architecture}}
\label{tab:gen}
\resizebox{0.95\linewidth}{!}{%
\begin{tabular}{@{}cccc@{}}
\toprule
\textbf{Layer Type} & \textbf{Kernels} & \textbf{Dimensions}            & \textbf{Activation} \\
\toprule
\rowcolor{Gray}
Convolution         & 128              & (\# input channels +  1, 1)    & Linear              \\

Convolution         & 128              & (\# input channels / 2 + 1, 1) & ELU                 \\

\rowcolor{Gray}
Upsampling          & -                & (scale factor – 1, 1)          & -                   \\

Convolution         & 128              & (\# input channels + 1, 1)     & ELU                 \\

\rowcolor{Gray}
Convolution         & 128              & (\# input channels / 2 + 1, 1) & ELU                 \\

Concatenate         & -                & -                              & -                   \\

\rowcolor{Gray}
Convolution         & 256              & (\# input channels / 2 + 1, 1) & ELU                 \\

Concatenate         & -                & -                              & -                   \\

\rowcolor{Gray}
Convolution         & 512              & (\# input channels / 2 + 1, 3) & ELU                 \\

Concatenate         & -                & -                              & -                   \\

\rowcolor{Gray}
Convolution         & 1                & (\# input channels + 1, 1)     & None                \\
\bottomrule

\end{tabular}
}
\end{table}

Our proposed WGAN model for EEG SR also consists of a generator and a discriminator. The generator architecture consists of the sequence of layers detailed in Fig. 1 with the parameters detailed in Table~\ref{tab:gen}. Similarly to~\cite{arjovsky2017wasserstein} we adopt a modified sequence of convolutional layers, which allows EEG data to be processed by Convolutional Neural Networks (CNNs) due to correlations across channels. This sequence is composed of convolutional layers with kernel dimensions that find the relationships between channels, (n, 1), where n = (\# input channels + 1). All convolutional layers enforce the same zero-padding to keep the same dimensions throughout. This sequence is then fed into one dense block~\cite{huang2017densely} composed of 3 densely connected convolutional sequences, followed by a convolutional layer whose outputs are the super-resolved channels. Note that the upsampling layer and the first subsequent convolutional layer are unique to SR models for a scale factor of 4 as the input channels need to be upsampled by 3 channel-wise.

The discriminator follows a similar scheme to the generator architecture aside from a few key differences detailed in Table~\ref{tab:disc}. The 4th convolutional layer has a stride of (4, 4) and is fed into a fully-connected layer. The final output activation of the model is linear to comply with the WGAN framework.

As in~\cite{ledig2017photo}, the generator’s parameters are first initialized through training the network in a supervised manner to map the downsampled LR data to the HR counterparts using a mean-squared error (MSE) loss function. This was found to prevent converging to local minima. The generator is then inserted into the GAN training process to fine-tune the model and find more optimal parameters in comparison to only using a distance metric as a loss function.

\section{Results}
\subsection{Training \& Hyperparameters}

\begin{table}[ht!]
\centering
\caption{\textbf{Discriminator Model Architecture}}
\label{tab:disc}
\resizebox{0.95\linewidth}{!}{%
\begin{tabular}{@{}cccc@{}}
\toprule
\textbf{Layer Type} & \textbf{Kernels} & \textbf{Dimensions}            & \textbf{Activation} \\
\toprule

\rowcolor{Gray}
Convolution         & 64               & (\# input channels + 1, 1)     & Linear              \\

Convolution         & 64               & (1, 3)                         & ELU                 \\

\rowcolor{Gray}
Concatenate         & -                & -                              & -                   \\

Convolution         & 128              & (\# input channels / 2 + 1, 1) & ELU                 \\

\rowcolor{Gray}
Concatenate         & -                & -                              & -                   \\

Convolution         & 256              & (\# input channels / 4 + 1, 3) & ELU                 \\

\rowcolor{Gray}
Fully-Connected     & 128              & -                              & ELU                 \\

Fully-Connected     & 1                & -                              & Linear              \\

\bottomrule
\end{tabular}%
}
\end{table}

Dropout regularization~\cite{srivastava2014dropout} was applied at the output of every activation within both the generator and discriminator, excluding the output layers. The generator and discriminator had dropout rates of 0.1 and 0.25, respectively. The $\alpha$ parameter for all ELU~\cite{clevert2015fast} activations was set to 1. The Adam optimizer~\cite{kingma2014adam} was used throughout with learning rate $\alpha=10^{-4}$, $\beta_1=0.5$, and $\beta_2=0.9$. The generator network was first pre-trained using a MSE (L2) loss for 50 epochs with a mini-batch size of 64. All hyperparameters were tuned to optimize performance on the validation set.  

The pre-trained generator was then fine-tuned using the WGAN framework losses with a gradient penalty weight of 10. The GAN training ratio was set to 3, which updates the discriminator once for every 3 generator updates. In addition to the WGAN loss function, a modification was made to multiply the WGAN loss by a factor of $10^{-2}$ and add a MSE loss on the generator output. This was inspired by the feature matching procedure from~\cite{salimans2016improved}, which specifies additional objectives for the generator to prevent overtraining on the discriminator. Also from~\cite{salimans2016improved}, the label smoothing technique was incorporated to assist in avoiding convergence to local minima. 

An evaluation of the model outputs of the validation and test datasets is displayed in Table~\ref{tab:results}. The quantitative results are compared to a baseline of bicubic interpolated channel data. Both MSE and mean absolute error (MAE) between upsampled and true EEG signals were calculated.

\subsection{Dataset V Classification Super Resolution Performance}
To further evaluate the validity of the SR data, we investigated the performance of classifying the mental imagery classes using the SR data. Deep neural network (DNN) classifiers were trained for both the precomputed features of the HR and SR data using the precomputed feature class labels. The DNN classifier consisted of 5 dense layers with 512, 256, 128, 64, and 3 neurons per layer, respectively. All layers contained ReLU~\cite{nair2010rectified} activations excluding the output layer, which consisted of a softmax activation. The classifiers were trained using a categorical cross-entropy loss optimized by the ADAM optimizer with learning rate $\alpha=10^{-3}$, $\beta_1=0.9$, and $\beta_2=0.99$. The class predictions with multiple metrics for the DNNs trained using the ground truth HR data and SR data by WGAN are recorded in Table~\ref{tab:cls}. 

\begin{table}[ht!]
\centering
\caption{\textbf{Super Resolution Performance Results.} Our proposed WGAN EEG Spatial Upsampling method significantly outperforms a baseline of Bicubic Interpolation commonly used in EEG upsampling pipelines.}
\label{tab:results}
\resizebox{0.8\linewidth}{!}{%
\begin{tabular}{@{}cccccc@{}}
\toprule
\multirow{2}{*}{\textbf{Dataset}} & \multirow{2}{*}{\textbf{Scale}} & \multicolumn{2}{c}{\textbf{Bicubic}} & \multicolumn{2}{c}{\textbf{WGAN}} \\ \cmidrule(l){3-6} 
                      &   & \textbf{MSE} & \textbf{MAE} & \textbf{MSE}    & \textbf{MAE}   \\
\toprule
\multirow{2}{*}{Val}  & 2 & 3.71E7       & 3.89E3       & \textbf{2.01E3} & \textbf{24.38} \\
                      & 4 & 7.23E7       & 6.42E3       & \textbf{8.53E3} & \textbf{63.83} \\
\midrule
\multirow{2}{*}{Test} & 2 & 3.75E7       & 3.91E3       & \textbf{2.06E3} & \textbf{24.66} \\
                      & 4 & 7.30E7       & 6.45E3       & \textbf{8.68E3} & \textbf{64.39} \\
\bottomrule
\end{tabular}%
}
\end{table}

\section{Discussion}
On the topic of CNNs for EEG time-series data, we highlight below some of the important findings throughout our research. Feature scaling techniques besides standard normalization decreased model performance. With regards to convolutional layers, using kernel dimensions that contained weights for each channel in the input and output layers improved performance significantly over the standard kernel dimensions used for images, e.g., $3\times3$, $9\times9$. Implementing concatenation connections instead of residual connections, popular in ResNet~\cite{he2016deep} architectures used in many Super Resolution papers, offered improved performance using a lesser amount of layers. A Linear activation on the input layer followed by ELU activations on the subsequent layers outperformed other popular neural network activation function combinations. 

It was notably difficult and time-consuming to train GANs for EEG data. We observed after testing different variants of GAN that WGAN appeared to be more stable during training. Replacing MSE with MAE in all loss functions produced SR EEG signals which were smoothed and did not contain similar frequency domain statistics as the HR data. It can be concluded that the task of EEG SR is highly sensitive to the loss function components used during training. 

Observing the results in Table~\ref{tab:results}, compared to bicubic interpolation, WGAN achieved $10^4$ fold and $10^2$ fold reduction in MSE and MAE, respectively, demonstrating the remarkable improvement of our proposed WGAN method in simultaneously reconstructing numerous missing EEG signals at a high resolution. Judging from the results in Table~\ref{tab:cls} it can be observed that classification of SR data produces minimal loss of accuracy when compared to ground truth signals, less than 4\% and 9\% for scale factors of 2 and 4, respectively. 

\section{Conclusion}
Our results conclude that our WGAN methods significantly improved over bicubic interpolation for the Dataset V EEG signals. We conclude that SR EEG by GAN is a promising approach to improve the spatial resolution of low-density EEG headsets. However, we intend to expand our work to perform well across multiple datasets for different classification tasks. Considerations for further work also include using different distance metrics than MSE for assessing signal similarity, as well as using other recent variations of the GAN framework to compare results.  

\begin{table}[ht!]
\centering
\caption{\textbf{Classification Performance Results on the Berlin BCI Competition III, Dataset V test set.} We remove 16 (\textit{scale=2)} and 24 (\textit{scale=4)} of the dataset's 32 EEG channels and use our proposed WGAN spatial upsampling method to recreate the missing channels using the remaining channels. We then train and evaluate a classifier on the upsampled channels and show minimal loss in performance compared to using the original dataset channels.}
\label{tab:cls}
\resizebox{0.9\linewidth}{!}{%
\begin{tabular}{@{}ccccc@{}}
\toprule
\textbf{Scale} &
\textbf{Metric} &
\textbf{Class} &
\textbf{HR} &
\textbf{WGAN (Ours)} \\
\toprule
\multirow{7}{*}{2} & Accuracy                   & - & 87.75 & 83.88 \\ \cmidrule(l){2-5} 
                   & \multirow{3}{*}{Precision} & 2 & 86.65 & 82.24 \\
                   &                            & 3 & 87.54 & 82.84 \\
                   &                            & 7 & 88.77 & 86.12 \\ \cmidrule(l){2-5} 
                   & \multirow{3}{*}{Recall}    & 2 & 84.33 & 80.21 \\
                   &                            & 3 & 88.57 & 84.37 \\
                   &                            & 7 & 89.62 & 86.26 \\ \midrule
\multirow{7}{*}{4} & Accuracy                   & - & 87.75 & 82.00 \\ \cmidrule(l){2-5} 
                   & \multirow{3}{*}{Precision} & 2 & 86.65 & 80.21 \\
                   &                            & 3 & 87.54 & 82.03 \\
                   &                            & 7 & 88.77 & 83.28 \\ \cmidrule(l){2-5} 
                   & \multirow{3}{*}{Recall}    & 2 & 84.33 & 77.73 \\
                   &                            & 3 & 88.57 & 81.34 \\
                   &                            & 7 & 89.62 & 85.94 \\ \bottomrule
\end{tabular}%
}
\end{table}

\section*{\centering \large Acknowledgements}
This work was supported by the Army Research Laboratory Cognition and Neuroergonomics Collaborative Technology Alliance (CANCTA) under Cooperative Agreement Number W911NF-10-2-0022.

\bibliographystyle{IEEEbib}
\bibliography{refs}

\end{document}